\documentclass[letterpaper, 10 pt, conference]{ieeeconf}  

\IEEEoverridecommandlockouts                              

\usepackage{booktabs}  
\usepackage{multirow}  
\usepackage{graphicx}
\usepackage{epsfig} 
\usepackage{times} 
\usepackage{amsmath} 
\usepackage{amssymb}  
\usepackage[ruled,vlined,linesnumbered]{algorithm2e}
\usepackage{float} 
\usepackage{url}
\usepackage{xcolor}
\usepackage{caption}
\usepackage{subcaption}
\title{\LARGE \bf
Topological Motion Planning Diffusion: Generative Tangle-Free Path Planning for Tethered Robots in Obstacle-Rich Environments
}

\author{Yifu Tian$^{1}$, Xinhang Xu, Thien-Minh Nguyen, Muqing Cao$^{2}$}

\begin{document}

\maketitle
\thispagestyle{empty}
\pagestyle{empty}

\begin{abstract}
In extreme environments such as underwater exploration and post-disaster rescue, tethered robots require continuous navigation while avoiding cable entanglement. Traditional planners struggle in these lifelong planning scenarios due to topological unawareness, while topology-augmented graph-search methods face computational bottlenecks in obstacle-rich environments where the number of candidate topological classes increases. To address these challenges, we propose Topological Motion Planning Diffusion (TMPD), a novel generative planning framework that integrates lifelong topological memory. Instead of relying on sequential graph search, TMPD leverages a diffusion model to propose a multimodal front-end of kinematically feasible trajectory candidates across various homotopy classes. A tether-aware topological back-end then filters and optimizes these candidates by computing generalized winding numbers to evaluate their topological energy against the accumulated tether configuration. Benchmarking in obstacle-rich simulated environments demonstrates that TMPD achieves a collision-free reach of 100\% and a tangle-free rate of 97.0\%, outperforming traditional topological search and purely kinematic diffusion baselines in both geometric smoothness and computational efficiency. Simulation with realistic cable dynamics further validates the practicality of the proposed approach.
\end{abstract}

\section{INTRODUCTION}

In extreme environments such as underwater exploration and post-disaster rescue, robots are usually deployed with physical tethers to ensure a stable power supply and high-bandwidth communication. In such navigation scenarios, path planning extends beyond collision avoidance and local trajectory efficiency; it requires maintaining global topological consistency to ensure the robot's tether remains strictly tangle-free.

Most existing approaches adapt classical search- or optimization-based planners with explicit topological representations to detect and reject entangling states. In particular, graph-search methods augment discrete planning states with topological descriptors (e.g., $H$-signatures) to distinguish tether configurations across homotopy classes \cite{kim2014, neptune}. However, in obstacle-dense environments the number of feasible homotopy classes proliferates rapidly, leading to a combinatorial expansion of the augmented state space and prohibitive search cost. This issue is further exacerbated by the fine spatial discretization required to resolve narrow passages, making real-time deployment challenging.

Recently, diffusion models have excelled at modeling multimodal trajectory distributions. Approaches like Motion Planning Diffusion \cite{mpd} (MPD) leverage GPU parallelism to generate diverse, collision-free candidates in constant time. However, these raw trajectories remain topologically unaware during continuous navigation, as they completely lack memory of the executed cable configuration.
To bridge this gap, we propose Topological Motion Planning Diffusion (TMPD). Our framework decouples the complex lifelong navigation task into a parallel generative front-end and a topological filtering back-end. Instead of explicitly searching in the augmented spatial-topological space, TMPD leverages a thermodynamic-inspired sampling mechanic within the diffusion model to propose a rich set of trajectory candidates across various homotopy classes. A tether-aware evaluation back-end then screens these candidates by utilizing general winding numbers to quantify cable entanglement, ensuring global tether safety.

To the best of our knowledge, TMPD is the first generative planning framework that integrates lifelong topological memory to solve the continuous tangle-free navigation problem. The main contributions of this paper are summarized as follows:

\begin{itemize}
    \item We propose a decoupled generative planning framework that integrates lifelong execution memory to generate tangle-free trajectories.
    \item We introduce a thermodynamic-inspired guidance schedule as the front-end to actively explore diverse homotopy classes and prevent mode collapse.
    \item We design an optimized, lazy-evaluation topological back-end that efficiently extracts the safest global path via generalized winding numbers and geometric curve shortening.
    \item We validate TMPD through extensive benchmarking in obstacle-rich environments in realistic simulations.
\end{itemize}

\section{Related Work}
\label{sec:related_work}

\subsection{Tethered Robot Planning}
\label{sec:related_tether}
Tethered robot planning differs from standard motion planning because feasibility depends not only on robot--obstacle collisions but also on the evolving cable configuration: obstacle contacts and tautness induce history-dependent reachability limits and entanglement risks, so planners often couple geometric search with topological reasoning. 

A major line of work augments discrete search states with topological descriptors that distinguish the tether configurations, e.g., distance-based metrics \cite{Igarashi2011}, topological invariants such as the $H$-signature \cite{bhattacharya2012topological,kim2014path,neptune}, topological braids \cite{Cao-RSS-23,cao2025braid}, topology-informed heuristics for faster A* planning \cite{kim2015multiheuristic}, and additional constraints targeting self-entanglement prevention \cite{yang2023self}. These ideas extend to more complex settings including 3D tethered motion using neighborhood-augmented graphs \cite{TopoGeoDistinctNAG} and sloped 3D terrains \cite{THAMP3D}.
Alternative formulations include optimization-based trajectory generation that parameterizes the tether (e.g., as connected points) and optimizes subject to collision/tether constraints \cite{Martinez2021optimization}, and deep reinforcement learning-based methods \cite{TangleFreeTetherDRL}. 
Topology-augmented graph search can become computationally expensive in obstacle-dense environments where the number of distinct topological classes proliferates, leading to large augmented state spaces and slow online planning.
Furthermore, grid-based searches inherently produce jagged paths and requires further refinement.
Relatedly, winding number has been used to represent interaction patterns in crowd navigation \cite{WindingThrough}, but has not been applied in tethered robot navigation. 

\subsection{Diffusion-Based Motion Planning and Constraints}
Diffusion probabilistic models have recently emerged as a transformative framework for continuous trajectory generation. Seminal works, including Diffuser \cite{diffuser} and Diffusion Policy \cite{dp}, reframe robotic motion synthesis as a progressive denoising process. By operating directly within the continuous trajectory space, this generative paradigm avoids the jaggedness of grid-based planners and has been successfully adapted to various complex domains. Planners such as DiPPeR \cite{dipper}, DiPPeST \cite{dippest}, LDP \cite{ldp}, and PDM \cite{pdm} utilize diffusion to propose smooth, kinodynamically feasible paths, effectively acting as intelligent sampling strategies.

While diffusion models excel at unconstrained generation, deploying them in dense obstacle fields necessitates adherence to geometric safety. Consequently, recent efforts in \textit{constrained diffusion} focus on aligning models with hard problem constraints. Techniques such as Classifier-Free Guidance (CFG) \cite{cfg}, Constraint-Guided Diffusion (CGD) \cite{cgd}, PGDM \cite{pgdm}, and SafeDiffuser \cite{safed, efficient, aligning} directly intervene in the denoising steps to steer the generative process toward feasible sub-regions. 

A prominent example in robotic navigation is Motion Planning Diffusion (MPD) \cite{mpd}, which guides inference using differentiable motion planning costs (e.g., collision and smoothness gradients). However, this strict enforcement of local safety introduces \textit{Topological Mode Collapse}. By heavily amplifying collision guidance weights to prevent any intersection, these frameworks suppress the inherent Langevin stochasticity of the diffusion process. Consequently, the generated samples tend to converge into the nearest, locally optimal spatial corridor. This restricts the diversity of the generated paths, making the planner prone to sub-optimal local minima and failing to explore alternative safe homotopy classes.

To address this limitation, our work TMPD modifies the guidance strategy. By leveraging the injected noise as a thermodynamic mechanism to escape local geometric constraints, the generative front-end proposes a topologically diverse candidate pool. Drawing inspiration from topological invariants \cite{neptune, kim2014}, the tether-aware topological back-end then evaluates and filters these candidates using a topological energy metric. Our framework unites the kinematic smoothness of generative models with the strict topological safety required for continuous navigation tasks.

\section{PRELIMINARIES}
In this section, we discuss the topology concepts and the standard formulation of Denoising Diffusion Probabilistic Models (DDPM), which serve as the core network architecture of our diffusion-based tangle-free path planning framework.
\subsection{Homotopy and Topological Winding Number}
We consider the workspace $\mathcal{W} \subset \mathbb{R}^2$ as a connected and bounded region. The environment contains a set of arbitrary-shaped obstacles $\mathcal{O} = \{O_1, O_2, \dots, O_m\}$. The obstacle-free configuration space is defined as $\mathcal{W}_{free} = \mathcal{W} \setminus \bigcup_{i=1}^m O_i$. Two curves, $\tau_1$ and $\tau_2$, sharing the exact same start and goal points, are considered \textit{homotopic} if one can continuously deform into the other strictly within $\mathcal{W}_{free}$. The set of all such mutually deformable curves forms a \textit{homotopy class}. For a tethered robot, its trajectory's homotopy class uniquely identifies the geometric state of the taut cable.

However, explicitly resolving homotopy equivalence is computationally expensive for real-time applications. Here we introduce the \textit{generalized topological winding number}. 

\begin{figure}[t]
    \centering
    \includegraphics[width=1.0\columnwidth]{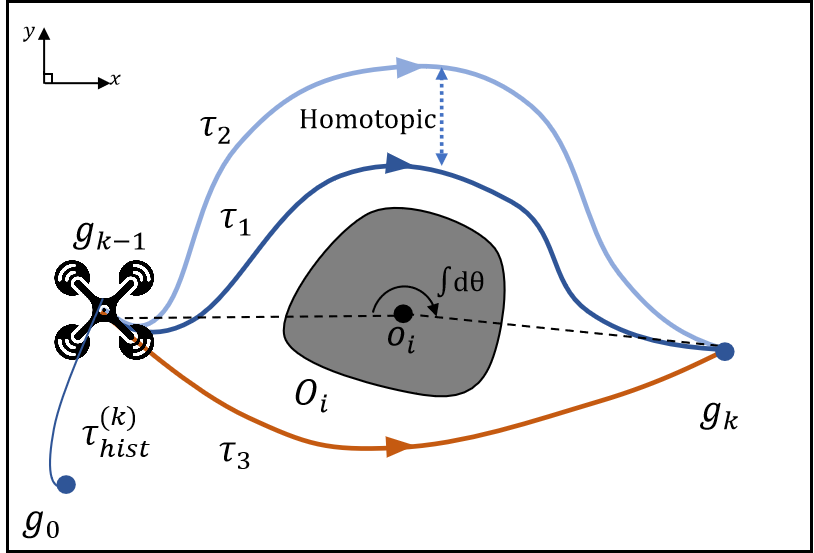} 
    \caption{Paths $\tau_1$ and $\tau_2$ (blue) belong to the same homotopy class as they can be continuously deformed into each other within the obstacle-free space. Path $\tau_3$ (orange) represents a distinct homotopy class, circumventing $O_i$ from the opposite side. The generalized winding number evaluates the accumulated angle $\int d\theta$ along the path relative to the obstacle center $o_i$.}
    \label{fig:winding_illustration}
\end{figure}

For a continuous planar open curve $C$ and an obstacle center $o_i = (x_{o_i}, y_{o_i})$ not lying on $C$, the fractional winding number computes the normalized accumulated angle that the curve travels counterclockwise around the reference point, defined by the line integral:
\begin{equation}
    W(C, o_i) = \frac{1}{2\pi} \int_{C} \frac{(x-x_{o_i})dy - (y-y_{o_i})dx}{(x-x_{o_i})^2 + (y-y_{o_i})^2}.
    \label{eq:winding_number}
\end{equation}
As illustrated in Fig. \ref{fig:winding_illustration}, winding number measures the extent to which the robot's cable wraps around a specific obstacle. For example, $|W(C, o_i)| \ge 1.0$ indicates a complete $360^{\circ}$ wrap around the pivot, representing an entanglement.

\subsection{Denoising Diffusion Probabilistic Models}
DDPMs are a class of generative models that learn to approximate a target data distribution $q(\mathbf{x}_0)$ by reversing a Markovian forward noise-addition process. The forward process gradually corrupts a data sample $\mathbf{x}_0$ into standard Gaussian noise $\mathbf{x}_T \sim \mathcal{N}(0, \mathbf{I})$ over $T$ discrete diffusion steps. The generative reverse process aims to recover the clean data by sampling from a learned Gaussian transition $p_\theta(\mathbf{x}_{t-1}|\mathbf{x}_t) = \mathcal{N}(\mu_\theta(\mathbf{x}_t, t), \Sigma_\theta(\mathbf{x}_t, t))$, where the mean $\mu_\theta$ is parameterized by a neural network (typically a U-Net). 

In our work, the state $\mathbf{x}$ represents a continuous trajectory $\tau$. The generative process is conditioned on the specific planning context $\mathbf{c}$ (e.g., start and goal states), yielding the conditional transition $p_\theta(\tau^{t-1}|\tau^t, \mathbf{c})$. The standard reverse sampling can be modified to incorporate external constraints without retraining the neural network. By injecting the gradients of a differentiable cost function $\mathcal{J}(\tau)$, the sampling manifold can be explicitly directed towards safe or optimal sub-regions. The guided reverse step at iteration $t$ can be approximated as:
\begin{equation}
    \tau^{t-1} = \mu_\theta(\tau^t, t, \mathbf{c}) - \alpha_t \nabla_{\tau} \mathcal{J}(\mu_\theta) + \sigma_t \mathbf{z}
    \label{eq:guided_sampling}
\end{equation}
where $\alpha_t$ scales the gradient guidance, $\sigma_t \mathbf{z}$ (with $\mathbf{z} \sim \mathcal{N}(0, \mathbf{I})$) represents the injected Langevin noise, and $\nabla_{\tau} \mathcal{J}$ pulls the trajectory away from high-cost regions (e.g., obstacles). 

This guided stochastic formulation naturally encapsulates an exploration-exploitation mechanism: the gradient term exploits local optimums to satisfy constraints, while the injected noise preserves the generative multimodality.

\section{Problem Formulation}
\label{sec:problem_formulation}
In our work, we address the lifelong tethered navigation problem. The robot is connected to a fixed anchor base $g_0 \in \mathcal{W}_{free}$ and sequentially navigate through a series of $N$ waypoints $G = \{g_1, g_2, \dots, g_N\} \subset \mathcal{W}_{free}$. Let $\tau^{(k)}$ denote the local trajectory segment planned at the $k$-th step, connecting $g_{k-1}$ to $g_k$. 

\begin{figure}[t]
    \centering
    \includegraphics[width=1.5\columnwidth]{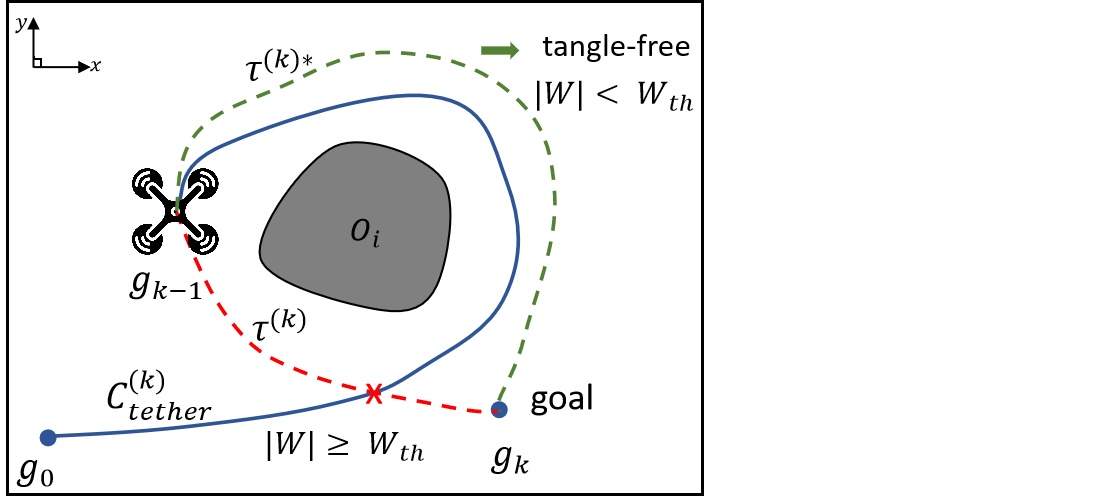}
    \caption{At the $k$-th step, the robot at $g_{k-1}$ is constrained by the history trajectory $\tau_{hist}^{(k)}$ (blue solid line) anchored at $g_0$. A kinematically feasible but topologically naive segment $\tau^{(k)}$ (red dashed line) heading directly to $g_k$ causes the global trajectory $\tau_{global}^{(k)}$ to entangle around obstacle $O_i$, violating the threshold $|W| < W_{th}$. $\tau^{(k)*}$ (green dashed line) is a valid tangle-free trajectory.}
    \label{fig:problem_formulation}
\end{figure}

Assuming the tether cannot penetrate obstacles and is subject to tension, it smoothly slides and tightens within the workspace. The actual spatial configuration of the tether at step $k$, denoted as $\mathcal{C}_{tether}^{(k)}$, remains a continuous deformation of the path traversed by the robot. This implies that the physical tether and the historical executed path share the exact same \textit{homotopy class}:
\begin{equation}
    \mathcal{C}_{tether}^{(k)} \simeq \tau_{hist}^{(k)},
\end{equation}
where $\tau_{hist}^{(k)}$ is the sequential spatial union of all previously executed segments: $\tau_{hist}^{(k)} = \tau^{(1)} \cup \tau^{(2)} \cup \dots \cup \tau^{(k-1)}$. This equivalence allows us to represent the entanglement state of the tether purely using the robot's kinematic execution history. Consequently, the total topological footprint induced by a newly proposed segment $\tau^{(k)}$ must be evaluated on the global concatenated trajectory:
\begin{equation}
    \tau_{global}^{(k)} = \tau_{hist}^{(k)} \cup \tau^{(k)}
\end{equation}

As illustrated in Fig.~\ref{fig:problem_formulation}, the core challenge in lifelong navigation is that a locally optimal, collision-free segment $\tau^{(k)}$ may cause the global trajectory $\tau_{global}^{(k)}$ to entangle around an obstacle $O_i \in \mathcal{O}$. Here we define $\text{Hom}(\tau)$ as the \textit{taut configuration} of a given trajectory $\tau$. Let $W(\cdot, O_i)$ be the fractional winding number operator defined in Eq.(\ref{eq:winding_number}). Formally, the continuous navigation process is defined as \textit{tangle-free} if and only if the global trajectory satisfies the following hard topological constraint:
\begin{equation}
    \forall O_i \in \mathcal{O}, \quad \left| W(\text{Hom}(\tau_{global}^{(k)}), O_i) \right| < W_{th}
\end{equation}
where $W_{th} \in (0, 1]$ denotes the critical winding threshold. In practice, a complete physical wrap corresponds to a mathematical value of $1.0$. In our work, we adopt a safety margin of $W_{th} = 0.95$ to prevent tension singularities at obstacle corners. Consequently, at each navigation step $k$, given the current state $g_{k-1}$, the target state $g_k$, and the execution history $\tau_{hist}^{(k)}$, the objective is to find an optimal continuous trajectory segment $\tau^{(k)*}$ that minimizes a composite objective function composed of global topological energy $\mathcal{J}_{t}$, kinematic smoothness $\mathcal{J}_{s}$, and path length $\mathcal{J}_{l}$:
\begin{equation}
    \tau^{(k)*} = \arg\min_{\tau^{(k)}} \Big( \mathcal{J}_{t}(\tau_{g}^{(k)}) + \lambda_{s} \mathcal{J}_{s}(\tau^{(k)}) + \lambda_{l} \mathcal{J}_{l}(\tau^{(k)}) \Big),
\end{equation}
subject to the constraints:

\textbf{1) Collision-Free Constraint:} $\tau^{(k)} \subset \mathcal{W}_{free}$.

\textbf{2) Tangle-Free Constraint:}
    \[
        \max\limits_{O_i \in \mathcal{O}} \left| W(\text{Hom}(\tau_{hist}^{(k)} \cup \tau^{(k)}), O_i) \right| < W_{th}.
    \]

\section{Methodology}
\label{sec:methodology}
Our method is inspired by the baseline paper \cite{mpd}. We propose Topological Motion Planning Diffusion (TMPD) by decoupling the task into a generative front-end and a topological filtering back-end.

\subsection{Generative Front-end via Diffusion}
\label{subsec:diffusion_prior}

Let a local trajectory segment at the $k$-th navigation step be parameterized as a sequence of $H$ waypoints $\tau^{(k)} = \{q_1, \dots, q_H\}$. MPD formulates motion planning as a conditional denoising process. To avoid notation collision, we denote the discrete diffusion timesteps using subscripts. Starting from pure Gaussian noise $\tau_T^{(k)} \sim \mathcal{N}(\mathbf{0}, \mathbf{I})$, where $T$ is the total number of diffusion steps, the model iteratively denoises it into a kinematically feasible path $\tau_0^{(k)}$ using a learned reverse process conditioned on boundary states $\boldsymbol{c} = \{g_{k-1}, g_k\}$:
\begin{equation}
    p_\theta(\tau_{0:T}^{(k)} | \boldsymbol{c}) = p(\tau_T^{(k)}) \prod_{t=1}^T p_\theta(\tau_{t-1}^{(k)} | \tau_t^{(k)}, \boldsymbol{c})
\end{equation}

To bias the generated samples toward obstacle-free and smooth regions without requiring expensive retraining, inference is guided by a composite cost functional. Building upon the generalized guided sampling step introduced in Eq. (\ref{eq:guided_sampling}), we define the guidance cost $\mathcal{J}(\tau)$ as a weighted sum of specific differentiable motion planning costs $c_i(\tau)$. The practical guided sampling step at iteration $t$ expands to:
\begin{equation}
    \tau_{t-1}^{(k)} = \mu_\theta(\tau_t^{(k)}, t, \boldsymbol{c}) - \alpha_t \sum_i \lambda_i \nabla_{\tau} c_i(\mu_\theta) + \sigma_t \mathbf{z},
    \label{eq:practical_guidance}
\end{equation}
where $\mu_\theta$ is the predicted mean from the neural network, $\lambda_i$ are the respective cost weights, $\alpha_t$ acts as the gradient scale, and $\mathbf{z} \sim \mathcal{N}(\mathbf{0}, \mathbf{I})$ is the injected Langevin noise.

In original diffusion-based planners, the guidance gradients $\nabla c_i$ often act as a dominant external potential field that `over-cools' the system. This causes the generated samples to prematurely collapse into the nearest local potential well and severely degrades the topological diversity of the candidate pool. 
To overcome this, we treat the trajectory sampling in TMPD through the lens of \textit{stochastic thermodynamics}. Our strategy is not to naively alter the gradient weights $\lambda_i$, but to schedule the sampling mechanics to preserve the system's entropy during critical stages. Specifically, we balance exploration and exploitation via the following hyperparameters:

\textbf{1) Stochasticity Injection ($\sigma_{extra}$):} We maintain a higher `temperature' by amplifying the Langevin noise scale. By explicitly injecting greater variance ($\sigma_{extra}$) into the generative process, we force the trajectories to explore wider spatial regions. 

\textbf{2) Guidance Scheduling ($\tau_{guide}$ and $n_{guide}$):} We implement a delayed-intervention schedule. By postponing the guidance onset ($\tau_{guide}$) and limiting the active steps ($n_{guide}$), we allow the system to undergo a `high-temperature' free exploration phase. This ensures the generative prior captures the full spectrum of homotopy classes before the guidance gradients eventually `freeze' the candidates into strictly collision-free configurations.

\begin{figure}[t]
    \centering
    \includegraphics[width=1.35\linewidth]{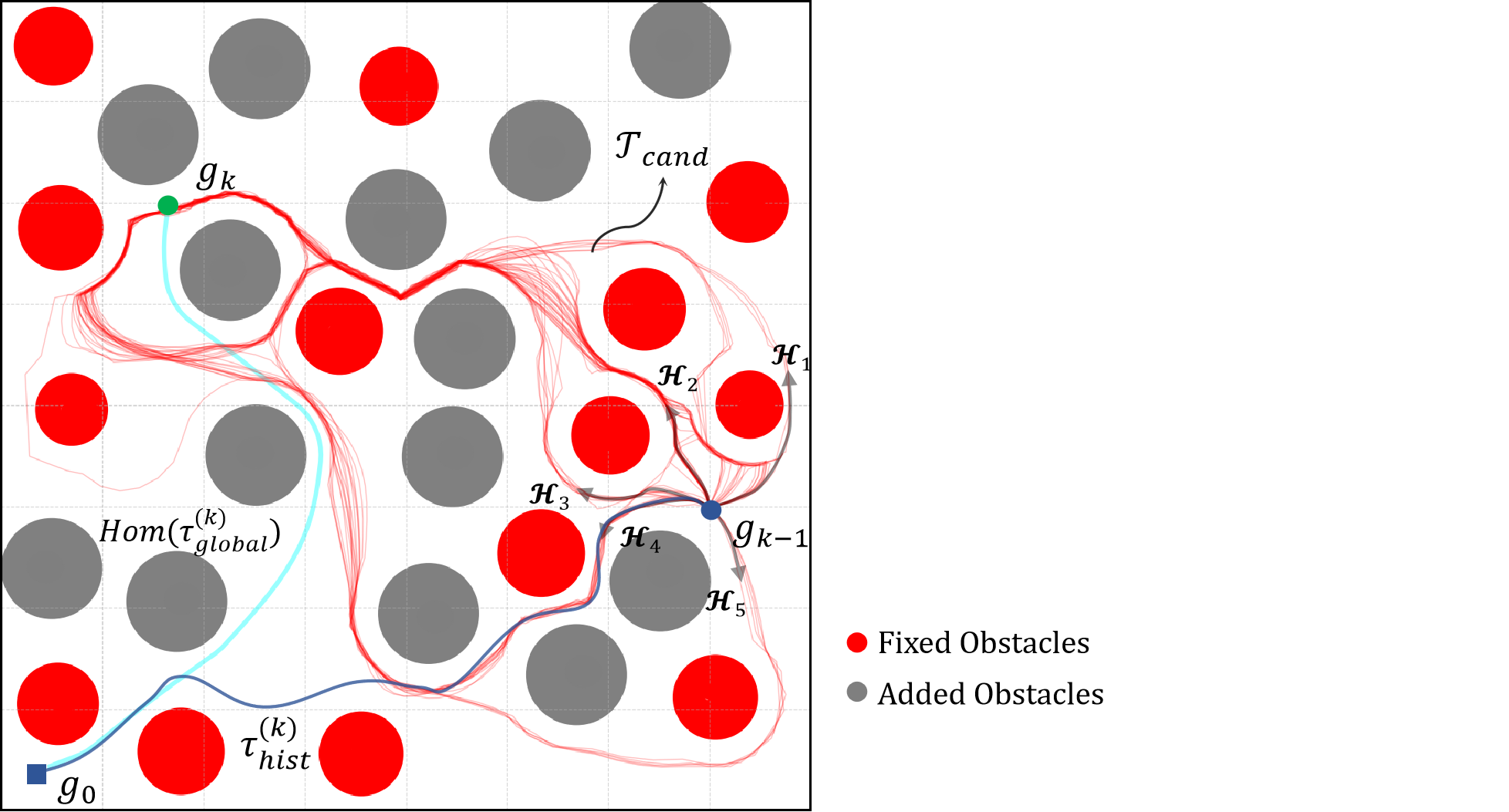}
    \caption{The robot currently resides at $g_{k-1}$, with the dark blue line representing the executed history trajectory $\tau_{hist}^{(k)}$ (i.e., the current physical cable configuration). The diffusion model generates a candidate pool $\mathcal{T}_{cand}$ (red lines) connecting $g_{k-1}$ to the target $g_k$, exploring diverse topological branches ($\mathcal{H}_1 \dots \mathcal{H}_5$). Tthe topological back-end then concatenates a specific candidate—in this case, a trajectory from the $\mathcal{H}_4$ branch—with the history. The cyan dashed line illustrates its resulting global taut configuration $\text{Hom}(\tau_{global}^{(k)})$.}
    \label{subsec:topo_backend}
\end{figure}

\textbf{3) Candidate Pool ($N_{samples}$):} To ensure the captured entropy is fully represented, we sample a batch of $B = N_{samples}$ trajectories simultaneously. By shifting the generative paradigm from finding a single path to maintaining a diverse "swarm" of candidates $\mathcal{T}_{cand} = \{\tau_1, \dots, \tau_{N_{samples}}\}$ (as visualized by the red trajectories in Fig. \ref{subsec:topo_backend}), TMPD reduces the mode collapse inherent in rigid guidance. We validate that this physics-inspired scheduling is essential for navigation with history-dependent constraints.
\subsection{Tether-Aware Topological Back-end}
\label{subsec:topo_posterior}

While the diffusion front-end proposes a diverse pool of candidate trajectories $\mathcal{T}_{cand}$, these kinematic paths are history-agnostic. To guarantee tether safety, our framework executes a three-stage `lazy-evaluation' topological back-end pipeline as shown in Algorithm \ref{alg:topo_selection}.

\begin{algorithm}[htbp]
\caption{Topological Back-end}
\label{alg:topo_selection}
\KwIn{Diffusion candidate pool $\mathcal{T}_{cand}$, history trajectory $\tau_{hist}^{(k)}$, obstacle set $\mathcal{O}$}
\KwOut{Optimal safe trajectory $\tau^{(k)*}$}

\tcp{Initialize set for unique homotopy candidates}
$\mathcal{M} \leftarrow \emptyset$\; 
\ForEach{candidate $\tau_i \in \mathcal{T}_{cand}$}{
    \If{$\tau_i$ forms a new homotopy class}{
        $\tau_{global} \leftarrow \tau_{hist}^{(k)} \oplus \tau_i$\;
        $\mathcal{J}_{topo} \leftarrow \text{EvalGlobalEnergy}(\tau_{global}, \mathcal{O})$\;
        $\mathcal{J}_{len} \leftarrow \text{PathLength}(\tau_i)$\;
        $\mathcal{J}_{total} \leftarrow \mathcal{J}_{topo} + \lambda \mathcal{J}_{len}$\;
        $\mathcal{M} \leftarrow \mathcal{M} \cup \{(\tau_i, \mathcal{J}_{total})\}$\;
    }
}

Sort $\mathcal{M}$ such that $\mathcal{J}_{total}^{(1)} \le \mathcal{J}_{total}^{(2)} \le \dots \le \mathcal{J}_{total}^{(|\mathcal{M}|)}$\;

\ForEach{ranked candidate $(\tau^{(m)}, \mathcal{J}_{total}^{(m)}) \in \mathcal{M}$}{
    $\tau_{global} \leftarrow \tau_{hist}^{(k)} \oplus \tau^{(m)}$\;
    \tcp{Curve Shortening}
    $\tau_{taut}^{(m)} \leftarrow \text{Hom}(\tau_{global})$\; 
    
    \tcp{Topological Veto}
    \If{$\forall O_j \in \mathcal{O}, |W(\tau_{taut}^{(m)}, O_j)| < W_{th}$}{
        \Return $\tau^{(m)}$\;
    }
}
\Return $\tau^{(1)}$\;
\end{algorithm}

We first group the raw candidates into a reduced set of \textit{homotopy equivalence classes} to avoid redundant evaluations. For each representative candidate $\tau_i$, evaluating its physical taut configuration is computationally expensive. Therefore, we compute \textit{global heuristic cost} $\mathcal{J}_{total}$ by concatenating it with the history $\tau_{hist}^{(k)}$. This composite function balances the global topological energy $\mathcal{J}_{topo}$ against the length of trajectories$\mathcal{J}_{len}$:
\begin{equation}
    \mathcal{J}_{total}(\tau_i) = \mathcal{J}_{topo}(\tau_{hist}^{(k)} \cup \tau_i) + \lambda \mathcal{J}_{len}(\tau_i),
\end{equation}
where $\mathcal{J}_{len}(\tau_i)$ is the Euclidean path length of the new segment. The global topological energy $\mathcal{J}_{topo}$ evaluates the cumulative winding number (via Eq. \ref{eq:winding_number}) directly on the raw concatenated trajectory:
\begin{equation}
    \mathcal{J}_{topo}(\tau_{global}) = \sum_{O_j \in \mathcal{O}} \Phi \Big( \big| W(\tau_{global}, O_j) \big| \Big),
\end{equation}
where $\Phi(x) = \alpha x^2$ is a quadratic barrier function penalizing aggressive entanglement. The clustered candidates are then sorted in ascending order based on this heuristic cost.

Then, instead of running geometric simplifications for all candidates, we iterate through the sorted list and perform a \textit{Lazy Geometric Evaluation}. For the ranked candidate $\tau_m$, we compute its \textit{taut configuration} $\text{Hom}(\tau_{hist}^{(k)} \cup \tau_m)$ using an iterative visibility-based \textbf{curve shortening} algorithm \cite{kim2014}. By pruning intermediate path segments that are mutually visible within $\mathcal{W}_{free}$, the trajectory collapses into a piecewise-linear shortest path. 

Finally, if the simplified taut cable satisfies the tangle-free constraint ($\forall O_j \in \mathcal{O}, |W| < W_{th}$), it is accepted as the optimal trajectory $\tau^{(k)*}$, and the search terminates.

\section{Experiments}
\label{sec:experiments}

To evaluate TMPD, our experiments are designed to answer three core questions: (1) How does our topological back-end enhance the baseline in single-query tasks? (2) How does TMPD benchmark against existing topology-aware planners in lifelong continuous navigation? (3) How do the diffusion hyperparameters affect the thermodynamic exploration-exploitation trade-off to prevent topological mode collapse? 

\subsection{Experimental Setup}
\textbf{Environments:} We evaluate our framework in bounded 2x2 spatial 2D environments characterized by randomly distributed obstacle fields. To test the planner's topological adaptability and robustness against unknown environments, a subset of the obstacles is randomly removed to open new potential homotopy classes for each trial. Up to 12 new spheres (with radii ranging from 0.08 to 0.11) and 12 new rectangular boxes (with edge lengths between 0.15 and 0.18) are added. To ensure the environment is intensely cluttered yet kinematically solvable, these additional obstacles are placed using rejection sampling to guarantee a minimum inter-obstacle clearance of 0.3.

\textbf{Baselines:} To ensure a fair comparison, we integrate our winding number entanglement detection into the classical planners to reject topologically invalid paths and benchmark TMPD against them. 
\begin{itemize}
    \item \textbf{Topology-Aware Classical Planners:} We develop \textit{Topo-A*} and \textit{Topo-RRT}. \textit{Topo-A*} represents a deterministic, grid-based heuristic search. Our implementation is similar to the topological planner in \cite{kim2014}. But instead of computing homotopy invariants, we use winding number to validate node expansions and augment the search space. \textit{Topo-RRT} represents randomized, sampling-based tree exploration. During the steering phase, it prunes any newly sampled branches if their accumulated winding number, combined with the execution history, violates the strict $W_{th}$ constraint.
    \item \textbf{Diffusion-based Path Planning Baseline:} We evaluate the original Motion Planning Diffusion (MPD) \cite{mpd}. It lacks a tether-aware topological back-end, serving as a critical ablation baseline to highlight the necessity of our method in lifelong navigation.
\end{itemize}

\textbf{Evaluation Metrics:} Over $N = 100$ independent trials, we record the following metrics: 
\begin{itemize}
    \item \textbf{Collision-free Reach:} The percentage of trials where the robot successfully reaches the goal state without any geometric collisions with the environment.
    \item \textbf{Tangle-Free Rate:} The percentage of successful trials where the absolute winding number of the taut tether remains below the critical threshold $W_{th}$.
    \item \textbf{Planning Time:} The average computational inference time per navigation step in seconds.
    \item \textbf{Path Length:} The total Euclidean distance of the traversed trajectory.
    \item \textbf{Topological Energy:} The accumulated topological cost $\mathcal{J}_{topo}$ to quantify the potential tension of partial wraps around obstacles.
    \item \textbf{Smoothness:} The kinematic smoothness of the generated trajectory is quantified by the cumulative $L_2$ norm of the discrete second-order derivatives (approximated accelerations) over the trajectory horizon. For non-smooth paths, this finite-difference formulation penalizes sharp corners, which manifest as instantaneous large velocity changes. 
\end{itemize}

\textbf{Dataset Generation and Training:}
We follow the data generation and training process established by MPD \cite{mpd}. For each environment, we offline-generate an expert dataset comprising 500 random start-goal contexts with 20 trajectories per context. These kinematically smooth and collision-free expert demonstrations are synthesized using a cascaded pipeline of RRTConnect, B-spline smoothing, and Stochastic-GPMP optimization. The diffusion model processes trajectories of dimension $H \times d$, where the planning horizon is set to $H=64$ and the state dimension is $d=2$ for planar navigation. Training is executed for 25 diffusion steps using an exponential noise schedule, utilizing a 5\% validation split for early stopping. To maximize computational efficiency, all simulation environments, generative inference processes, and the topological back-end operators are implemented in PyTorch \cite{pytorch} to leverage high-throughput GPU parallelization. All model training and experimental benchmarking were executed on a workstation equipped with an NVIDIA RTX A6000 GPU.
\subsection{Tether-Free Navigation}
Before evaluating lifelong navigation, we conduct a tether-free, point-to-point experiment to assess the baseline generative performance. Both MPD and TMPD are tasked to navigate through 100 dense environments.

\begin{figure}[t]
    \centering
        \includegraphics[trim={8cm 2cm 6cm 1cm}, clip, width=0.4\textwidth]{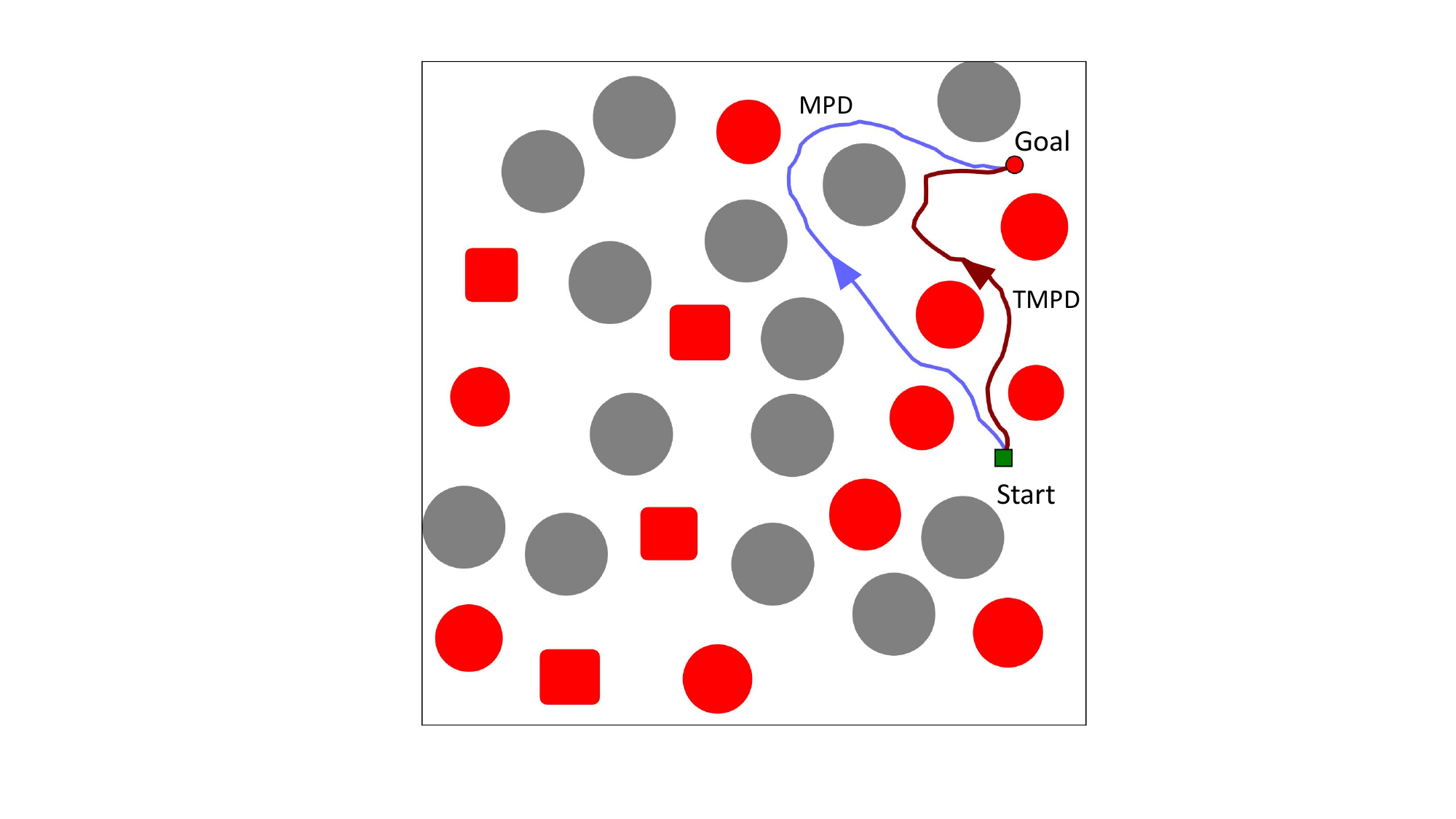} 
    \caption{While MPD (blue) often produces irregular trajectories that unnecessarily skirt obstacle boundaries, our TMPD (dark red) utilizes topological energy and kinematic penalties to recover optimal, smooth, and safe paths. }
    \label{fig:single_task_overlay}
\end{figure}

\begin{table}[htbp]
\centering

\setlength{\tabcolsep}{4pt} %
\resizebox{\columnwidth}{!}{
\small
\begin{tabular}{lccccc}
\toprule
\textbf{Method} & \textbf{Collision-free Reach} & \textbf{Topo-Energy} & \textbf{Length} & \textbf{Smoothness} & \textbf{Time} \\
\midrule
MPD & 100.0\% & $3.93 \pm 1.54$ & $1.68 \pm 0.37$ & $2544.11 \pm 662.21$ & $0.56s$ \\
TMPD (Ours) & 100.0\% & $3.67 \pm 1.35 $ & $1.72 \pm 0.41$ & $491.76 \pm 233.01$ & $1.03s$ \\
\bottomrule
\end{tabular}%
}
\caption{Results for 100 Trials}
\label{tab:single_task}
\end{table}

As shown in Table~\ref{tab:single_task}, while both methods achieve a 100\% collision-free reach rate, MPD outputs sub-optimal trajectories from the diffusion distribution. This results in geometric jaggedness and higher topological energy. By scoring and filtering candidates through our topological back-end, TMPD achieves a \textbf{5.2x improvement} in kinematic smoothness and reduces topological tension, at the cost of a 0.50-second overhead for back-end evaluation.

\subsection{Lifelong Tethered Navigation Benchmark}
For continuous-space methods (Topo-RRT, MPD, and TMPD), we set the robot collision radius to $r=0.05$. Topo-RRT is granted a generous exploration budget of $50,000$ iterations with a step size of $0.05$. To ensure a fair comparison within the same dense domain, for the grid-based Topo-A*, we compute at the resolution of $0.01$ m and relax its collision radius to $r=0.02$.
\begin{figure*}[htbp]
    \centering
    \includegraphics[trim={1.5cm 10cm 0cm 0cm}, clip, width=1.12\linewidth]{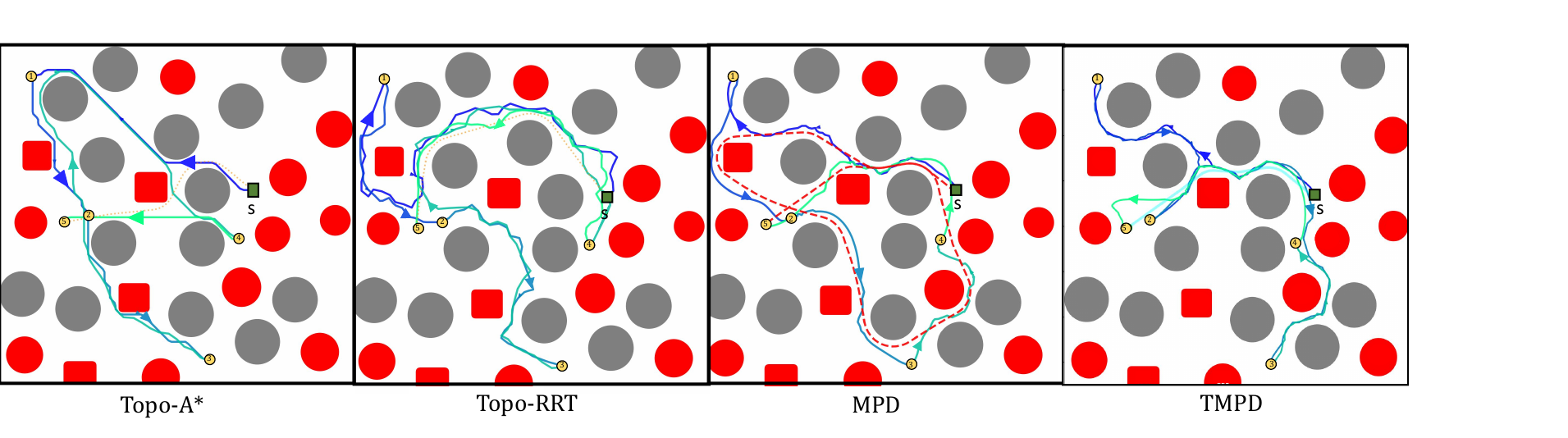}
    \caption{From left to right: Topo-A*, Topo-RRT, MPD, and our proposed TMPD. The grey circles represent randomly added obstacles, while the red squares denote fixed obstacles used during training. The blue lines indicate the execution history.}
    \label{fig:baseline_comparison_qualitative}
\end{figure*}

\begin{table*}[htbp]
\centering

\setlength{\tabcolsep}{5pt}
\begin{tabular}{@{}lcccccc@{}} 
\toprule
\textbf{Method} & \textbf{Collision-free Reach ($\uparrow$)} & \textbf{Tangle-free ($\uparrow$)} & \textbf{Time (s) ($\downarrow$)} & \textbf{Topo-Energy ($\downarrow$)} & \textbf{Path Length ($\downarrow$)} & \textbf{Smoothness ($\downarrow$)} \\ \midrule

Topo-RRT & $90.2\%$ & $86.0\%$ & $3.68 \pm 9.65$ & $4.54 \pm 2.38$ & $1.920 \pm 0.37$ & $535.46 \pm 132.19$ \\
Topo-A* & $98.0\%$ & $93.0\%$ & $2.01 \pm 1.94$ & $4.87 \pm 2.25$ & $ 1.279 \pm 0.18$ & $159.907 \pm 35.948$ \\
MPD      & $96.8\%$ & $40.0\%$ & $0.93 \pm 1.61$ & $5.93 \pm 2.17$ & $1.538 \pm 0.19$ & $343.87 \pm 72.55$ \\
\midrule
\textbf{TMPD (Ours)} & $\mathbf{100.0\%}$ & $97.0\%$ & $1.35 \pm 0.13$ & $4.74 \pm 2.44$ & $1.660 \pm 0.27$ & $344.63 \pm 76.70$ \\

\bottomrule
\multicolumn{7}{l}{\footnotesize \textit{Note: TMPD uses $\sigma_{extra}=0.8$, $\tau_{guide}=0.1$.}}
\end{tabular}
\caption{Comparison of TMPD against Baseline Planners in Lifelong Navigation}
\label{tab:baseline_comparison}
\end{table*}

As summarized in Table~\ref{tab:baseline_comparison}, classical search methods struggle with topological feasibility in obstacle-rich environments. Regarding the \textbf{Collision-Free Reach} metric, Topo-RRT and Topo-A* show noticeable performance degradation (90.2\% and 98.0\% respectively) and long planning times (e.g., Topo-RRT's large variance of ± 9.65s). When topological winding constraints are imposed on top of dense physical obstacles, Topo-RRT suffers from a compounded "narrow passage" problem, frequently failing to sample valid extensions. Conversely, the grid-based Topo-A* hits discretization bottlenecks, leading to exponential state-space explosion and timeouts due to many augmented topological states. For the \textbf{Tangle-Free Rate}, despite integrating winding number checks, classical baselines still show topological failures in some cases($86.0\%$ and $82.0\%$). As visualized in Fig.~\ref{fig:baseline_comparison_qualitative}, these planners often generate jerky (Topo-RRT) or overly tight (Topo-A*) paths that graze obstacles, dropping their tangle-free success to 86.0\% and 93.0\%, respectively. Furthermore, the MPD achieves a significantly low 40.0\% Tangle-free Rate (illustrated by the entangled red dashed trajectory in Fig.~\ref{fig:baseline_comparison_qualitative}), proving that history-agnostic diffusion models cannot maintain tether safety. In contrast, TMPD overcomes these limitations. It achieves a $100.0\%$ Collision-Free Rate and a $97.0\%$ Tangle-free Rate. Our framework decouples generative spatial exploration from topological filtering. This separation maintains a diverse pool of candidates. As a result, TMPD prevents the robot from falling into local topological traps. It ensures lifelong tether safety while maintaining a practical and stable inference time ($1.35s \pm 0.13$).

\subsection{Ablation Study}
\label{sec:ablation_tradeoff}

Here we ablate our scheduling parameters to evaluate their impact on topological safety and smoothness. Table~\ref{tab:mega_ablation} validates how our sampling mechanics balance this exploration-exploitation trade-off.

\begin{table*}[t]
\centering

\setlength{\tabcolsep}{8pt}
\begin{tabular}{@{}llccc@{}}
\toprule
\textbf{Hyperparameter} & \textbf{Value} & \textbf{Tangle-free ($\uparrow$)} & \textbf{Time (s) ($\downarrow$)} & \textbf{Smoothness ($\downarrow$)} \\ 
\midrule

\multirow{4}{*}{Noise Scale ($\sigma_{extra}$)} 
& $0.5$ & $89.0\%$ & $\mathbf{1.54 \pm 0.20}$ & $56.29 \pm 26.82$ \\
& $\mathbf{0.8}$ & $\mathbf{95.0\%}$ & $1.59 \pm 0.18$ & $54.66 \pm 25.62$ \\
& $1.8$ & $92.0\%$ & $2.35 \pm 0.50$ & $42.97 \pm 28.33$ \\
& $3.0$ & $83.0\%$ & $2.27 \pm 0.63$ & $\mathbf{24.47 \pm 27.62}$ \\
\midrule

\multirow{4}{*}{Candidate Pool ($N_{samples}$)} 
& $30$ & $93.0\%$ & $\mathbf{1.52 \pm 0.11}$ & $54.69 \pm 26.25$ \\
& $\mathbf{70}$ & $\mathbf{97.0\%}$ & $1.67 \pm 0.21$ & $54.87 \pm 25.42$ \\
& $100$ & $95.0\%$ & $1.76 \pm 0.17$ & $56.66 \pm 27.88$ \\
& $120$ & $95.0\%$ & $1.77 \pm 0.16$ & $56.72 \pm 27.50$ \\
\midrule

\multirow{3}{*}{Guidance Start ($\tau_{guide}$)} 
& $\mathbf{0.1}$ & $\mathbf{97.0\%}$ & $\mathbf{1.40 \pm 0.14}$ & $54.88 \pm 27.71$ \\
& $0.3$ & $96.0\%$ & $1.82 \pm 0.24$ & $\mathbf{53.33 \pm 25.75}$ \\
& $0.7$ & $96.0\%$ & $2.50 \pm 0.32$ & $54.81 \pm 26.92$ \\
\midrule

\multirow{3}{*}{Guidance Steps ($n_{guide}$)} 
& $5$ & $90.0\%$ & $\mathbf{1.14 \pm 0.29}$ & $\mathbf{26.54 \pm 9.78}$ \\
& $\mathbf{10}$  & $\mathbf{97.0\%}$ & $1.40 \pm 0.14$ & $54.88 \pm 27.29$ \\
& $15$ & $96.0\%$ & $1.76 \pm 0.14$ & $87.54 \pm 45.93$ \\
\bottomrule
\multicolumn{5}{l}{\footnotesize \textit{Note: Across all trials, Collision-free Reach Rate consistently maintained at 100\%.}}
\end{tabular}
\caption{Ablation Study of TMPD Hyperparameters}
\label{tab:mega_ablation}
\end{table*}
At a low noise scale ($\sigma_{extra}=0.5$), the process acts exploitatively, yielding a suboptimal Tangle-Free Rate (TFR) of 89.0\%. Increasing noise to $\sigma_{extra}=0.8$ allows trajectories to escape local minima into alternative homotopy classes (TFR 95.0\%), but excessive noise ($\sigma_{extra}=3.0$) overwhelms geometric guidance (TFR 83.0\%). To further expand the topological search, increasing parallel samples ($N_{samples}$) from 30 to 70 improves the TFR from 93.0\% to 97.0\% before plateauing against linear computational costs. Finally, balancing the geometric guidance steps ($n_{guide}=10$) is required: under-refinement ($n_{guide}=5$) fails to satisfy topological constraints (TFR 90.0\%), whereas over-constraining the model ($n_{guide}=15$) degrades kinematic smoothness. It is worth noting that TMPD can achieve a 100\% tangle-free rate by scaling the generative candidate pool ($N_{samples}$). However, expanding the sampling budget requires more computational expenses. Our selected configuration represents an optimal trade-off.
\subsection{Unity Simulation}
\begin{figure}[t]
    \centering
    \includegraphics[trim={9cm 2cm 5cm 4cm}, clip, width=1.2\linewidth]{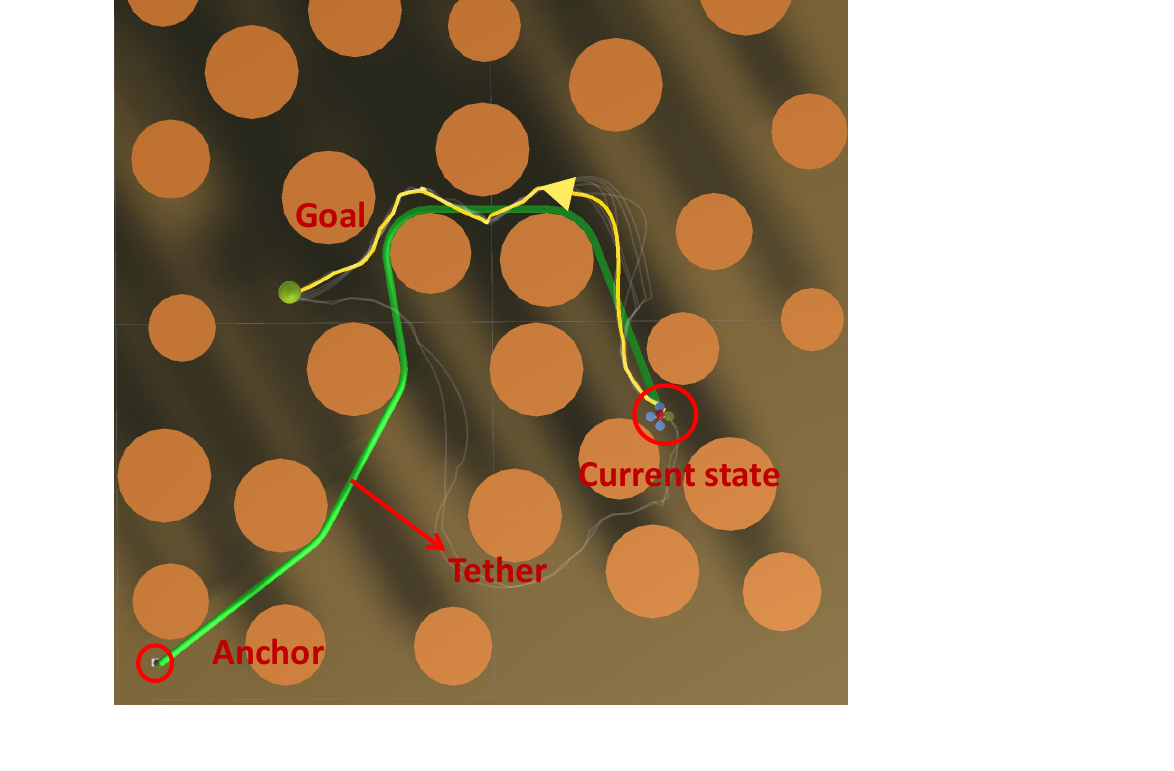}
    \caption{The physical tether (green) is in a semi-enclosed wrapped state. While the generative front-end proposes multiple spatial candidates (white), TMPD identifies the topologically optimal, tangle-free trajectory (yellow) to safely proceed without entanglement.}
    \label{fig:sim}
\end{figure}
To bridge the gap between kinematic path generation and real-world physical deployment, we deployed TMPD framework in Unity integrated with \textbf{AGX Dynamics for Unity}\footnote{\url{https://www.algoryx.se/agx-unity/}}, providing a more accurate cable simulation and mass-spring-based continuous collision detection framework.

As illustrated in Fig.~\ref{fig:sim}, the tether (solid green) has formed a semi-enclosed wrapping state around multiple obstacles due to the execution history. The generative front-end proposes a batch of trajectory candidates (visualized as white curves). Tether-agnostic planners would likely select a path that crosses the opposite side of the adjacent obstacles, leading to an entanglement. TMPD successfully identifies and extracts the optimal safe trajectory (highlighted in yellow).
\section{CONCLUSIONS}
In this paper, we presented Topological Motion Planning Diffusion (TMPD), a tether-aware diffusion-based path planning framework. By separating the continuous navigation task into a generative front-end and a tether-aware topological back-end, TMPD is successful in generating tangle-free paths on average $97\%$ in obstacle-rich environments. We validate in AGX Dynamics Unity environments and showcase its exceptional collision-free and tangle-free success rates. It provides a favorable trade-off between real-time inference performance and global topological safety.

Our future work will focus on reducing the time cost and extending the TMPD framework to 3D spatial environments, where tether entanglement requires more complex topological invariants. Additionally, we plan to deploy the system on physical hardware to investigate the sim-to-real transfer of our topological back-end under real-world sensing uncertainties and dynamic moving obstacles.




\bibliographystyle{IEEEtran} 

\bibliography{references}

\end{document}